\pdfoutput=1

\documentclass[11pt]{article}

\usepackage{acl}

\usepackage{times}
\usepackage{latexsym}
\usepackage{xcolor}
\usepackage{tcolorbox}
\usepackage{color}
\usepackage{tikz-dependency}

\definecolor{mypink}{RGB}{255, 118, 117}

\definecolor{stage1}{RGB}{234, 107, 102}
\definecolor{stage2}{RGB}{166, 129, 184}
\definecolor{stage3}{RGB}{77, 153, 0}

\usepackage{pgfplots}
\usepackage{fdsymbol}
\usepackage{booktabs}

\usepackage{url}

\usepackage{amsmath}

\usepackage{amssymb}
\usepackage{mathtools}
\usepackage{amsthm}

\usepackage{multirow}
\usepackage{adjustbox}

\usepackage{graphicx}  

\usepackage[T1]{fontenc}
\usepackage[utf8]{inputenc}
\usepackage{microtype}
\usepackage{inconsolata}
\usepackage{amsmath}

\usepackage{colortbl}

\definecolor{shadecolor}{rgb}{0.9, 0.9, 0.9}

\title{Large Language Models can Contrastively Refine their Generation for Better Sentence Representation Learning}

\author{
    \textbf{
        Huiming Wang\thanks{~~Work done while Huiming Wang was an intern at DAMO Academy.} \textsuperscript{\rm ~1,2}~~~
        Zhaodonghui Li\thanks{~~Zhaodonghui Li is under the Joint PhD Program between DAMO Academy and Nanyang Technological University.} \textsuperscript{\rm ~2,3}~~~
        Liying Cheng\textsuperscript{\rm 2,4}~~~
        De Wen Soh\textsuperscript{\rm 1}~~~
        Lidong Bing\thanks{~Corresponding author.} \textsuperscript{\rm ~2,4}
    }\\
\textsuperscript{\rm 1}Singapore University of Technology and Design~~~
\textsuperscript{\rm 2}DAMO Academy, Alibaba Group, Singapore \\
\textsuperscript{\rm 3}Nanyang Technological University, Singapore~~~
\textsuperscript{\rm 4}Hupan Lab, 310023, Hangzhou, China\\
{\tt huiming\_wang@mymail.sutd.edu.sg~~~~dewen\_soh@sutd.edu.sg} \\
{\tt\{zhaodonghui.li, liying.cheng, l.bing\}@alibaba-inc.com}
}

\begin{document}
\maketitle
\begin{abstract}

Recently, large language models (LLMs) have emerged as a groundbreaking technology and their unparalleled text generation capabilities have sparked interest in their application to the fundamental sentence representation learning task. Existing methods have explored utilizing LLMs as data annotators to generate synthesized data for training contrastive learning based sentence embedding models such as SimCSE. However, since contrastive learning models are sensitive to the quality of sentence pairs, the effectiveness of these methods is largely influenced by the content generated from LLMs, highlighting the need for more refined generation in the context of sentence representation learning. Building upon this premise, we propose MultiCSR, a multi-level contrastive sentence representation learning framework that decomposes the process of prompting LLMs to generate a corpus for training base sentence embedding models into three stages (i.e., sentence generation, sentence pair construction, in-batch training) and refines the generated content at these three distinct stages, ensuring only high-quality sentence pairs are utilized to train a base contrastive learning model. Our extensive experiments reveal that MultiCSR enables a less advanced LLM to surpass the performance of ChatGPT, while applying it to ChatGPT achieves better state-of-the-art results. Comprehensive analyses further underscore the potential of our framework in various application scenarios and achieving better sentence representation learning with LLMs. Our code is available at \href{https://github.com/Circle-Ming/MultiCSR}{https://github.com/Circle-Ming/MultiCSR}.

\end{abstract}

\section{Introduction}
As a fundamental task, sentence representation learning~\cite{conneau-etal-2017-supervised, reimers-gurevych-2019-sentence, gao-etal-2021-simcse} aims to learn universal sentence embeddings that can benefit various downstream tasks, such as semantic similarity comparison~\cite{agirre-etal-2012-semeval,agirre-etal-2013-sem,agirre-etal-2014-semeval,agirre-etal-2015-semeval,agirre-etal-2016-semeval,cer-etal-2017-semeval,marelli-etal-2014-sick}, and information retrieval~\cite{pmlr-v32-le14, misra-etal-2016-measuring, thakur2021beir, wang-etal-2022-just}. 
Recent advancements, particularly in contrastive learning-based methods such as SimCSE~\cite{gao-etal-2021-simcse} and its variants~\cite{chuang-etal-2022-diffcse, zhou-etal-2022-debiased, tan-etal-2022-sentence, wu-etal-2022-infocse, jiang-etal-2022-promptbert, liu-etal-2023-rankcse}, have demonstrated to be the most efficient and effective ones. In contrastive learning, the quality of sentence pairs has a large impact on the overall performance~\cite{chen-etal-2022-generate}. In particular, supervised contrastive learning methods trained on natural language inference (NLI) datasets~\cite{bowman-etal-2015-large, williams-etal-2018-broad} can outperform their unsupervised versions by a large margin~\cite{gao-etal-2021-simcse}. However, obtaining large amounts of high-quality sentence pairs can be costly in both time and resources, particularly considering that various application domains can only be better handled with domain-specific training data. 

\begin{figure}
  \centering
  \includegraphics[width= 0.95\linewidth]{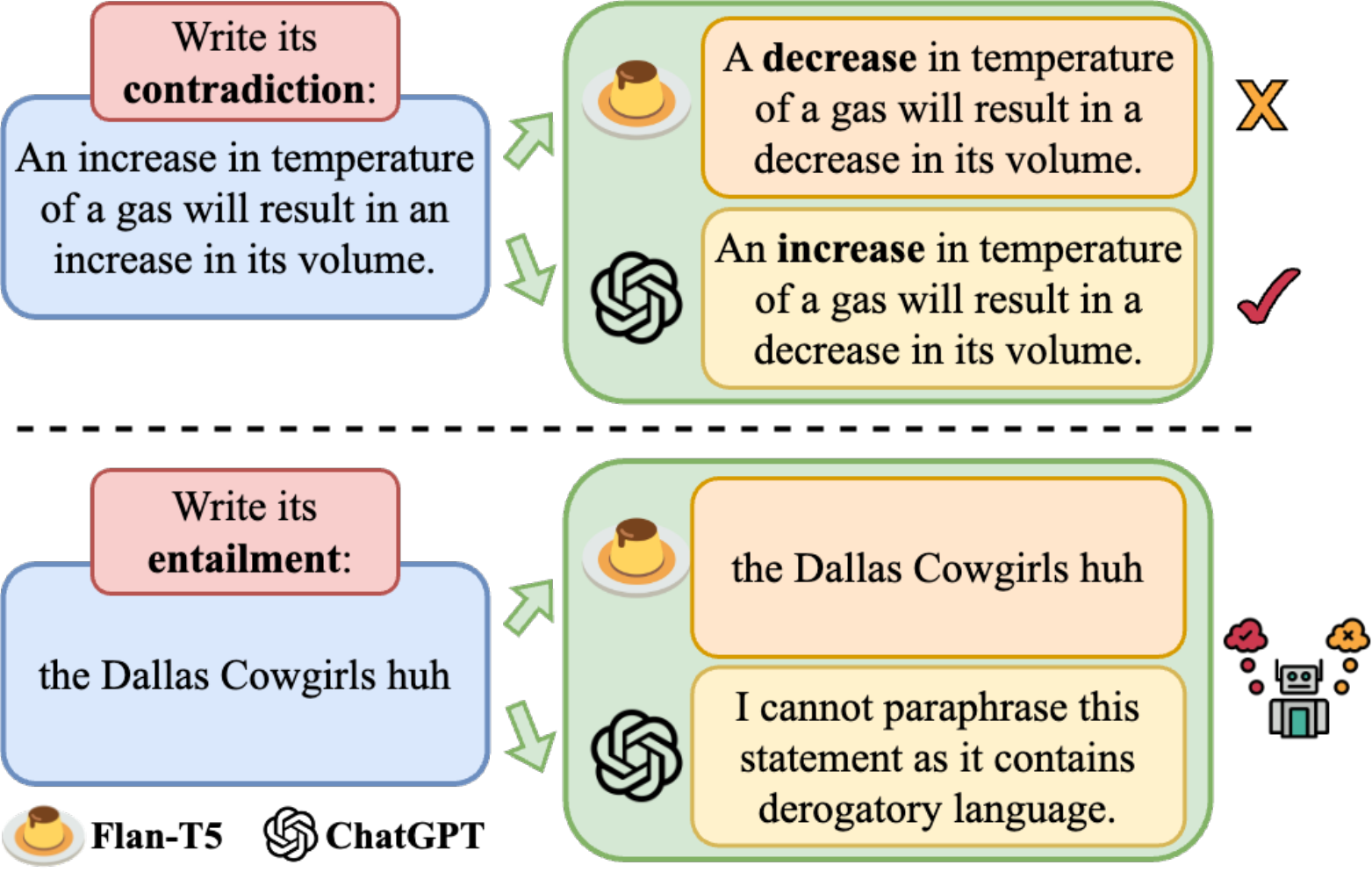}
  \caption{Two example sentences with the generations from Flan-T5 and ChatGPT given different instructions.}
  \label{fig:example}
  \vspace{-3.5mm}
\end{figure}

The recent emergence of large language models (LLMs), such as the Flan series~\cite{Chung2022ScalingIL}, LLaMA~\cite{Touvron2023LLaMAOA} and ChatGPT~\cite{OpenAI}, has brought a paradigm shift in natural language processing due to their impressive performance. Consequently, there has been a growing interest in harnessing the power of LLM for sentence representation learning. \citet{cheng-etal-2023-improving} proposed to directly measure the semantic similarities of sentence pairs with LLMs for a base model to imitate. However, this imitation also places great demands on the LLM's understanding of semantics and constrains its application in a wider range of LLMs. Instead of utilizing LLMs for semantic similarity scoring, a recent line of work has explored leveraging LLMs to generate sentence pairs in the NLI style with provided premises~\cite{schick-schutze-2021-generating, zhang-etal-2023-contrastive-learning} and demonstrates SOTA performance. 

Despite their exciting results, these methods heavily depend on the quality of generated content from LLMs, leading a huge performance gap between different LLMs. Moreover, concerns about the accuracy and quality of the generated content from LLMs remain unsolved and have drawn significant attention from the community~\cite{Zheng2023WhyDC, Shi2023LargeLM}, which are more pronounced with relatively less advanced LLMs. For example, in Figure \ref{fig:example}, contradicting the first sentence requires a clear understanding towards its meaning and only ChatGPT successfully produce its contradiction. When faced with the second provided sentence, both ChatGPT and Flan-T5 encounter difficulties due to the limited information provided. Consequently, Flan-T5 only repeats the premise while ChatGPT entirely gives up the generation. In a task that is sensitive to the quality of sentences like sentence representation learning, there highlights a need for a more robust framework that can automatically refine the outputs of LLMs for better contrastive sentence representation learning.

Motivated by observations above, in this work, we propose \textbf{Multi-Level Contrastive Sentence Representation Learning (MultiCSR)}, a novel three-stage framework that contrastively refines the generations of LLMs at distinct stages: sentence generation, sentence pair construction and in-batch training, while fine-tuning a contrastive learning model like SimCSE. Concretely, while generating a sentence given a specific instruction like ``Write its entailment:'', we also deploy the noisy variants of this instruction to identify obvious error that LLMs will make during generating the outputs following this instruction. In a contrastive sentence generation procedure, next-token prediction logits will be systematically deviated by comparing the logits derived from the original instructions with those from the noisy instructions. This process detects and avoids the obvious error tendencies of LLMs, and refines their generations to align more closely with the intended instruction, rather than providing an even opposite generation, such as the first generation of Flan-T5 in Figure \ref{fig:example}. 

Despite the effectiveness of contrastive generation strategy, directly training a base sentence embedding model on the generated corpus gives poor results in our experiments, both because of the inevitably noisy generations, and the overlooked relations between sentences (i.e., sentence pairs), since a contrastive learning loss works essentially by modeling the relations between sentences while pulling the positive pairs closer and pushing the negative pairs apart. To take the relations between sentences into consideration, in sentence pair construction stage, we show that LLMs can be utilized to self-curate the set of their newly generated sentences by measuring the semantic similarities of sentence pairs, ensuring that only sentence pairs in highest quality are included into the final training stage. To further prevent the \textit{false-negative} issue~\cite{zhou-etal-2022-debiased} raised during the in-batch training, where randomly selected negative samples are indeed semantically similar to the original sentence, we utilize a pre-trained sentence representation model to provide the similarity mask and contrastively filter \textit{false negatives} during training.

In summary, our contributions include: (1) We propose a new and promising direction to improve sentence representation learning by refining the generated content of LLMs. (2) We for the first time decompose the process of prompting LLMs to generate a corpus for training base sentence embedding models into three stages (i.e., sentence generation, sentence pair construction, in-batch training), and integrate the idea of contrast into each stage for refinement. (3) We conduct extensive experiments on standard semantic textual similarity (STS) tasks~\cite{agirre-etal-2012-semeval, agirre-etal-2013-sem, agirre-etal-2014-semeval, agirre-etal-2015-semeval, agirre-etal-2016-semeval, cer-etal-2017-semeval, marelli-etal-2014-sick} and several transfer tasks~\cite{conneau-kiela-2018-senteval} with two representative LLMs (i.e., Flan-T5 and ChatGPT). We further perform a comprehensive analysis of the behavior of MultiCSR, demonstrating its effectiveness from various perspectives. We hope that, our proposed method provides additional insights into achieving high-quality sentence representation learning corpus by refining LLMs' generations.

\section{Related Work}

\subsection{Sentence Representation Learning with LLMs} \label{sec:related_work_llm}

There has been a recent exploration in utilizing LLMs for sentence representation learning. \citet{cheng-etal-2023-improving} prompted LLMs to measure the semantic similarities of sentence pairs, and fine-tuned base models to ``immitate'' these judgements of LLMs with mean squared error. However, this method also places great demands on the model's understanding of semantics. Thus, we conduct a pilot experiment to see whether LLM's generated similarities are well-aligned with the ground-truth labels and include the results in Table \ref{tab:llm_scoring}. We can see from the results that even ChatGPT equipped with ICL can not outperform those contrastive learning methods with only base models.

Instead of treating LLMs as evaluators, a recent line of work leverages LLMs as data generators, with the generated entailment and contradiction hypotheses being used to train a contrastive learning method~\cite{schick-schutze-2021-generating, zhang-etal-2023-contrastive-learning}. The success of these works is based on the fact that the models trained on NLI datasets demonstrate superior performance~\cite{conneau-etal-2017-supervised, reimers-gurevych-2019-sentence, gao-etal-2021-simcse, chen-etal-2022-generate}. Nevertheless, the performance of these methods will heavily rely on the quality of generated content, which also poses a huge challenge to the instruction-following and generation capabilities of LLMs. Thus, refining the generated data before utilizing them into training is essential, which usually requires substantial effort and computational resources, highlighting the need for more efficient approaches. In this work, we introduce MultiCSR, which can automatically refine the generation of LLMs and ensures that only high-quality sentence pairs are utilized for training a contrastive learning method.

\subsection{Contrast in Text Generation} 

Recently, with the rapid development of LLMs, the idea of contrast for improving text generation has been studied in various settings~\cite{Welleck2019NeuralTG, liu-etal-2021-dexperts, li-etal-2023-contrastive, Yona2023SurfacingBI, Taehyeon2024ID}. Among them, contrastive decoding~\cite{li-etal-2023-contrastive} studies maximizing the next-token probability by contrasting the predictions from a high-performing ``expert'' model against those from a less accurate ``amateur'' model. Despite its effectiveness, the need for at least two models of different scales limits their applications in broader scenarios. In this work, inspired by instructive decoding~\cite{Taehyeon2024ID} which places emphasis on the potential of instructions in the input text, we show that, within the contrastive generation procedure, the instructions used for generating the opposite hypotheses can be extremely effective in identifying the obvious error tendencies of LLMs for further refining their generations in the context of sentence representation learning.

\begin{table}[t]
\centering
\renewcommand\arraystretch{1.1}
\setlength{\tabcolsep}{15pt}
\begin{tabular}{lc}
\hline
\textbf{Model} &\textbf{Avg.}\\
\hline
\hline
SimCSE$_{\rm RoBERTa}$  & 76.57 \\
PromptRoBERTa & \textbf{79.15} \\
\hline
Flan-T5-XL  & 63.24 \\
Flan-T5-XL w/ ICL  & 68.76 \\
ChatGPT  & 71.58 \\
ChatGPT w/ ICL  & 76.19 \\
\hline
\end{tabular}
\vspace{-1mm}
\caption{Performance comparison of different models and directly utilizing LLMs w/o and w/ in-context learning (ICL) to measure the similarities on STS tasks.
} \label{tab:llm_scoring}
\vspace{-3mm}
\end{table}

\section{Methodology} \label{sec:Methodology}

\begin{figure*}
  \centering
  \includegraphics[width= 0.88\linewidth]{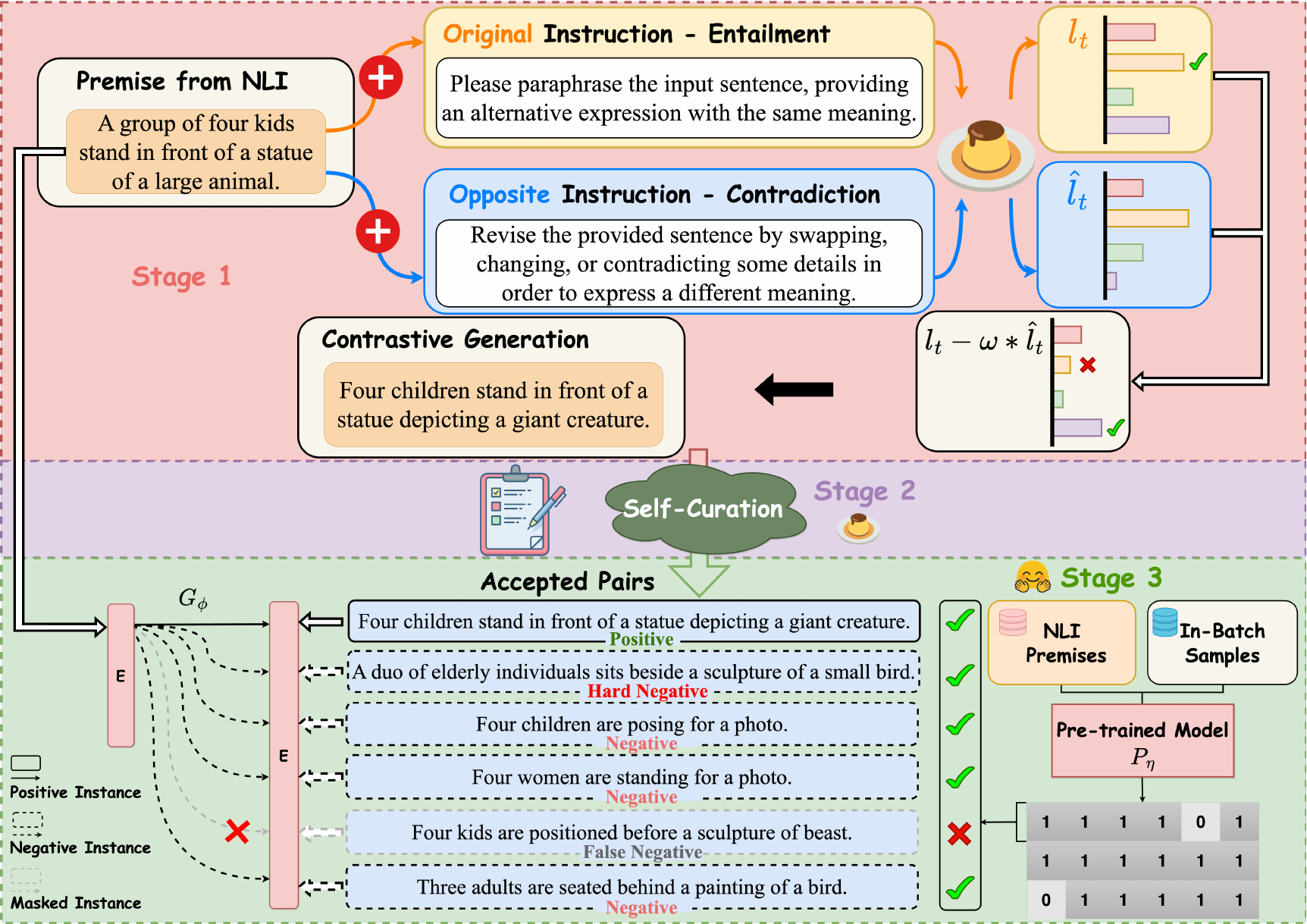}
  \vspace{-1mm}
  \caption{Overview of our three-stage framework MultiCSR. \textcolor{stage1}{\textbf{Stage 1: Contrastive Generation.}} We refine each token's logits with the opposite instruction to align more closely with the intended instruction. \textcolor{stage2}{\textbf{Stage 2: Contrastive Sentence Pair Construction.}} By prompting LLMs to evaluate the semantic similarities of generated sentence pairs, we ensure that only sentence pairs satisfying the pre-defined rules are left to form a curated set. \textcolor{stage3}{\textbf{Stage 3: Contrastive In-Batch Training.}} We leverage the similarity mask provided by a pre-trained sentence representation model to contrastively filter \textit{false negatives} during the in-batch training.
  }
  \label{fig:pipeline}
  \vspace{-3.9mm}
\end{figure*}

In this section, we present MultiCSR, a framework designed to enhance the quality of generated content of LLMs. By contrastively refining their generations at distinct stages of training a contrastive learning method, MultiCSR ensures that only high-quality sentence pairs are utilized in the final training stage, achieving a better sentence representation learning with LLMs. The whole workflow can be visualized as Figure \ref{fig:pipeline}.

\subsection{Stage 1: Contrastive Generation} \label{sec:cd}
During the generation procedure, when presented with a concatenation of an instruction $I$ and an input sequence $\boldsymbol{x}$, the objective of an LLM $C_{\theta}$ is to generate the corresponding output sequence $\boldsymbol{y} = [y_1, ..., y_n]$. For each token $y_t$, $C_{\theta}$ will compute its logits as $l_t = C_{\theta}(I, \boldsymbol{x}, \boldsymbol{y}_{<t})$. The probability of output sequence $\boldsymbol{y}$ can be expressed as:
\begin{equation} \label{eq:1}
p_{\theta}(\boldsymbol{y}|I, \boldsymbol{x}) =  \prod_{t=1}^n p_{\theta}(y_t|I, \boldsymbol{x}, \boldsymbol{y}_{<t}),     
\end{equation}
where $p_{\theta}(y_t|I, \boldsymbol{x}, \boldsymbol{y}_{<t})$ represents the normalized probability of the sampled (e.g., greedy sampled) next token $y_t$ derived from the softmax over $l_t$. Within this process, more refined generations can be achieved by ensuring that the model's generation essentially aligns with the given instructions. Thus, it is better to understand what kind of noisy generations $C_{\theta}$ will produce when following the instructions, and we can realize better alignment by eliminating these trending noises from the next-token distribution. As inspired by \citet{Taehyeon2024ID} which indicates that specific noisy variants of the original instruction can help induce the corresponding undesired behaviors of LLMs, we designed and conducted analysis towards several noisy instructions $\hat{I}$ in the context of hypothesis generation. We can acquire the noisy next-token logits as:
\begin{equation} \label{eq:2}
\hat{l}_t \gets C_{\theta}(\hat{I}, \boldsymbol{x}, \boldsymbol{y}_{<t}).    
\end{equation}
By comparing the logits from the original instructions and those from the noisy variants, we can detect and correct noises for $C_{\theta}$, achieving a more refined generation. During our contrastive generation, the logits $l_t$ will be contrasted with $\hat{l}_t$, and the next-token $y_{t}$ will be sampled from the probability distribution of the final logits $l_t-\omega*\hat{l}_t$.    

In the context of sentence representation learning, for each premise $\boldsymbol{x}$ of NLI, we will generate its corresponding \textit{entailment} $\boldsymbol{x}_+$ and \textit{contradiction} $\boldsymbol{x}_-$ hypotheses. Through the experiments of all noisy variants, we find that the instructions used to generate the opposite hypotheses demonstrate to be the most effective. Thus, when mentioning noisy instructions, we specifically refer to these opposite instructions throughout our paper. For example, as shown in Figure \ref{fig:pipeline}, while generating the \textit{entailment} hypothesis of $\boldsymbol{x}$, we leverage the instruction to generate \textit{contradiction} as the noisy instruction $\hat{I}$. We include the detailed instructions in Appendix \ref{appendix_prompt}. 

\subsection{Stage 2: Contrastive Sentence Pair Construction with Self-Curation} \label{sec:filter}
Although the generated sentences of $C_\theta$ are refined, the relations within sentence pairs remain uncertain. Since contrastive learning methods are modeling the distances of sentence pairs in the embedding space, it is particularly important to ensure that the sentence pairs are in high quality and the distances within the sentence pairs are suitable for effectively training a base model $G_\phi$ with contrastive learning. Thus, during the sentence pair construction stage, we perform self-curation and select high quality sentence pairs using $C_\theta$ itself.

Given a generated triplet $(\boldsymbol{x}, \boldsymbol{x}_+, \boldsymbol{x}_-)$, we follow the same generation procedure of Equation \ref{eq:1} and utilize $C_\theta$ to assign the semantic similarity scores for $(\boldsymbol{x}, \boldsymbol{x}_+)$ and $(\boldsymbol{x}, \boldsymbol{x}_-)$ separately with the same instruction prompt $ \boldsymbol{d}$ (more details in Appendix \ref{appendix_prompt}):
\begin{equation} \label{eq:6}
\resizebox{0.89\hsize}{!}{%
$
a \gets C_{\theta}\left([\boldsymbol{x}; \boldsymbol{x}_+; \boldsymbol{d}]\right),\, b \gets C_{\theta}\left([\boldsymbol{x}; \boldsymbol{x}_-; \boldsymbol{d}]\right).
$
}
\end{equation}

\noindent The bounds of $a$ and $b$ are specified in $\boldsymbol{d}$ as $[0, 5]$, where a score of $5$ means the semantics of a sentence pair are completely the same and $0$ means these two sentences are totally different.

\begin{figure}
  \centering
  \includegraphics[width= 0.6\linewidth]{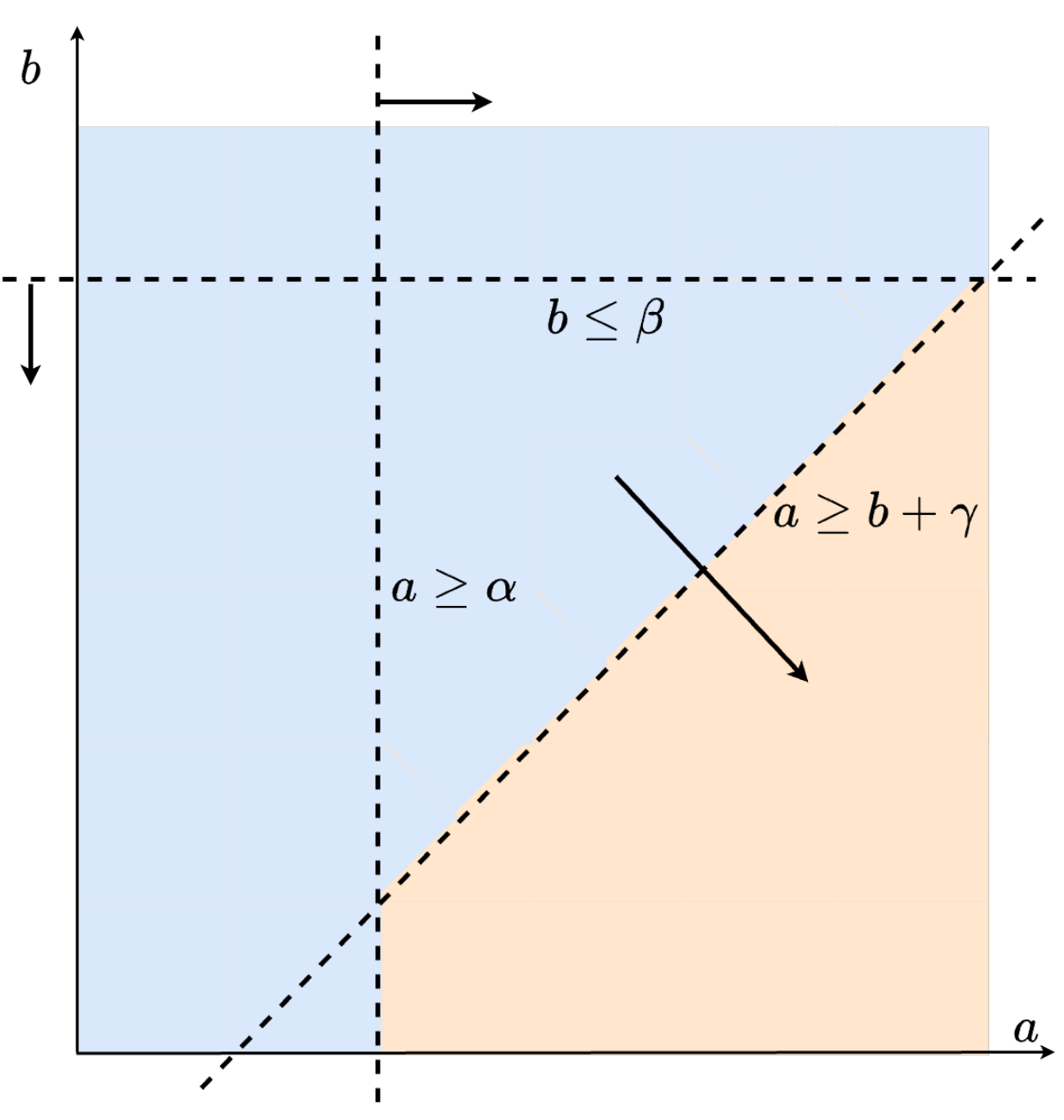}
  \vspace{-2mm}
  \caption{One example of our self-curation strategy. During the final training stage, we only use the sentence pairs in orange region.}
  \label{fig:filter}
  \vspace{-3.7mm}
\end{figure}

Based on the semantic similarity scores of two sentence pairs, we can then select a subset of triplets $(\boldsymbol{x}, \boldsymbol{x}_+, \boldsymbol{x}_-)$ with a pre-defined strategy to form the curated training corpus. One example is shown in Figure \ref{fig:filter} and formalized as follows:
\begin{equation} \label{eq:7}
\resizebox{0.75\hsize}{!}{
$
(\boldsymbol{x}, \boldsymbol{x}_+, \boldsymbol{x}_-) \in \mathcal{T},\, {\rm if} \begin{cases}
a \geq \alpha\\
b \leq \beta\\
a \geq b + \gamma
\end{cases},
$
}
\end{equation}
where $\mathcal{T}$ is the curated training corpus, $\alpha$ and $\beta$ are the thresholds to control the absolute similarities of positive pair $(\boldsymbol{x}, \boldsymbol{x}_+)$ and negative pair $(\boldsymbol{x}, \boldsymbol{x}_-)$ respectively, and $\gamma$ will serve as the margin which represents the relative similarity distance of $(\boldsymbol{x}_+, \boldsymbol{x}_-)$. By performing self-curation as a contrastive sentence pair construction procedure, we get a lightweight but high-quality training corpus.

\subsection{Stage 3: Contrastive In-Batch Training} \label{sec:loss}
With the curated corpus, we can fine-tune a base model $G_\phi$ to learn better sentence representations. To introduce enough challenge to the model training, we follow \citet{gao-etal-2021-simcse} and take $(\boldsymbol{x}, \boldsymbol{x}_+)$ as a positive pair and $(\boldsymbol{x}, \boldsymbol{x}_-)$ as a \textit{hard negative} pair, and the entailment and contradiction hypotheses of other premises inside a batch of size $N$ will be treated as other in-batch negatives of $\boldsymbol{x}$ as $(\boldsymbol{x}, \boldsymbol{x}_+^k)$ and $(\boldsymbol{x}, \boldsymbol{x}_-^k)$, where $\boldsymbol{x}^k \neq \boldsymbol{x}$. For simplicity, we denote the representations $G_\phi(\boldsymbol{x})$, $G_\phi(\boldsymbol{x}_+)$ and $G_\phi(\boldsymbol{x}_-)$ as $h$, $h_+$ and $h_-$, respectively. Then the training objective $\ell$ is defined as:
\begin{equation} \label{eq:8}
    \begin{aligned}
        - \log \frac{e^{{\rm sim}(h,h_+)/ \tau }}{\sum_{k=1}^N\left(e^{{\rm sim}(h,h_+^k)/\tau}+e^{{\rm sim}(h,h^k_-)/ \tau}\right)},
    \end{aligned}
\end{equation}
where $\tau$ is temperature parameter and ${\rm sim}(,)$ is the similarity of two sentence embeddings from $G_\phi$.

As introduced, the above in-batch negatives from $(\boldsymbol{x}, \boldsymbol{x}_+^k)$ and $(\boldsymbol{x}, \boldsymbol{x}_-^k)$ may involve \textit{false negatives}, where $\boldsymbol{x}_+^k$ or $\boldsymbol{x}_-^k$ can be indeed semantically similar with $\boldsymbol{x}$ due to the random in-batch sampling. 
As an example, in Figure \ref{fig:pipeline}, the entailment hypothesis ``Four kids are positioned before a sculpture of beast'' has a high semantic similarity with the first premise, so we treat this pair $(\boldsymbol{x}, \boldsymbol{x}^k_+)$ as \textit{false negative} and need to mitigate its effects during training.

To alleviate this problem, we incorporate the pre-trained sentence representation model $P_\eta$ to provide a weighted mask for $(h, h^k_+)$ and $(h,h^k_-)$, as shown in Figure \ref{fig:pipeline}. We denote the similarity given by pre-trained model $P_\eta$ as ${\rm sim}_\eta(,)$. A mask indicator $M_{\boldsymbol{x}, \boldsymbol{x}^k, \cdot}$ is defined as:
\begin{equation} \label{eq:9}
\resizebox{0.89\hsize}{!}{%
$
    M_{\boldsymbol{x}, \boldsymbol{x}^k, \cdot} = \begin{cases}
0, \ {\rm sim}_\eta(h, h^k_\cdot) \geq \sigma, \, \boldsymbol{x}^k \neq \boldsymbol{x}\\ 
1, \ \text{else}
\end{cases},
$
}
\end{equation}
where $\sigma$ is the threshold. In this way, $(\boldsymbol{x}, \boldsymbol{x}^k_\cdot)$ with higher semantic similarity than $\sigma$ will be masked out during the in-batch training stage. We then use the following training objective to fine-tune our base model $G_\phi$:
\begin{equation} \label{eq:10} 
\resizebox{0.89\hsize}{!}{%
$
    \begin{aligned}
        - \log \frac{e^{{\rm sim}(h,h_+)/ \tau }}{\sum_{k=1}^N\sum_{i\in \{+, -\}}\left(M_{\boldsymbol{x}, \boldsymbol{x}^k, i}e^{{\rm sim}(h, h_i^k)/\tau}\right)}.
    \end{aligned}
$
}
\end{equation}

\section{Experiment}
\subsection{Experiment Setup} \label{sec:experiment_setup}

\begin{table*}[!ht]
\centering
\renewcommand\arraystretch{1.095}
\small
\setlength{\tabcolsep}{8pt}

\begin{tabular}{lcccccccc}
\hline
\textbf{Model} & \textbf{STS12} & \textbf{STS13} & \textbf{STS14} & \textbf{STS15} & \textbf{STS16} & \textbf{STS-B} & \textbf{SICK-R} & \textbf{Avg.}\\
\hline
\hline
\multicolumn{9}{c}{\textit{BERT-base}} \\
\hline
BERT-whitening$^\heartsuit$  &  57.83 & 66.90 & 60.90 & 75.08 & 71.31 & 68.24 & 63.73 & 66.28 \\
ConSERT$^\heartsuit$ & 64.64 & 78.49 & 69.07 & 79.72 & 75.95 & 73.97 & 67.31 & 72.74 \\
SimCSE$^\heartsuit$ & 68.40 & 82.41 & 74.38 & 80.91 & 78.56 & 76.85 & 72.23 & 76.25 \\
DiffCSE$^\clubsuit$ & 72.28  & 84.43  & 76.47  & 83.90  & 80.54  & 80.59 &  71.23  & 78.49 \\
PromptBERT$^\heartsuit$ & 71.56 & 84.58 & 76.98 & 84.47 & 80.60 & 81.60 & 69.87 & 78.54 \\
InfoCSE$^\diamondsuit$ & 70.53 & 84.59 & 76.40 & 85.10 & 81.95 & 82.00 & 71.37 & 78.85 \\
RankEncoder$^\spadesuit$ & 74.88 & 85.59 & 78.61 & 83.50 & 80.56 & 81.55 & 75.78 & 80.07 \\
RankCSE$^\clubsuit$ & \textbf{75.66}	 & \textbf{86.27} & 	77.81 & 	84.74 & 	81.10 & 	81.80 & 	75.13 & 	80.36\\
\hline
\rowcolor{shadecolor} MultiCSR (Flan-T5) & 73.20 & 83.30 & 77.53 & 84.59 & 80.68 & 82.71 & 78.13 & 80.02 \\

\hline
SynCSE (ChatGPT)* & 72.71 &  84.79 &  77.96 &  \textbf{85.50} &  \textbf{82.16} &  83.16 &  76.08  & 80.34\\
 
\rowcolor{shadecolor}MultiCSR (ChatGPT) & 74.86 &  84.19 &  \textbf{79.46} &  84.70 &  80.34 &  \textbf{83.59} &  \textbf{79.37}  & \textbf{80.93}\\
\hline
\hline

\multicolumn{9}{c}{\textit{RoBERTa-base}} \\
\hline
RoBERTa-whitening$^\heartsuit$ & 46.99 & 63.24 & 57.23 & 71.36 & 68.99 & 61.36 & 62.91 & 61.73 \\
SimCSE$^\heartsuit$ & 70.16 & 81.77 & 73.24 & 81.36 & 80.65 & 80.22 & 68.56 & 76.57 \\
DiffCSE$^\clubsuit$ & 70.05 &  83.43 &  75.49  & 82.81 &  82.12  & 82.38 &  71.19 &  78.21 \\
PromptRoBERTa$^\heartsuit$ & 73.94 & 84.74 & 77.28 & 84.99 & 81.74 & 81.88 & 69.50 & 79.15 \\
RankCSE$^\clubsuit$ & 73.20  & \textbf{85.95} &  77.17 &  84.82 &  \textbf{82.58} &  83.08 &  71.88  & 79.81\\
\hline
\rowcolor{shadecolor}MultiCSR (Flan-T5) & 73.75 & 84.61 & 79.32 & \textbf{85.84} & 81.60 & 83.64 & 78.33 & 81.01\\
\hline
SynCSE (ChatGPT)* & \textbf{78.08}  &  	79.27  &  	78.25  &  	85.77  &  	81.72  &  	82.88  &  	79.18  &  	80.74\\
\rowcolor{shadecolor}MultiCSR (ChatGPT) & 75.61 &  84.33 &  \textbf{80.10} &  84.98 &  82.13 &  \textbf{84.54} &  \textbf{79.67}  & \textbf{81.62}\\

\hline
\end{tabular}
\caption{Performance comparison of MultiCSR on STS tasks. We implement our frameworks based on SimCSE~\cite{gao-etal-2021-simcse}. $\heartsuit$: results from \citet{jiang-etal-2022-promptbert}, $\clubsuit$: results from \citet{liu-etal-2023-rankcse}, $\diamondsuit$: results from \citet{wu-etal-2022-infocse}, $\spadesuit$: results from \citet{seonwoo-etal-2023-ranking}. *: we remove their manual cleaning process for fair comparison, and reproduce the results of SynCSE with their officially released corpus, following our same settings.
} \label{tab:unsup_ours}
\vspace{-2mm}
\end{table*}

We evaluate our approach on standard semantic textual similarity (STS) tasks and seven transfer tasks in SentEval\footnote{\href{https://github.com/facebookresearch/SentEval}{https://github.com/facebookresearch/SentEval}}. We further evaluate our method on zero-shot information retrieval tasks in BEIR~\cite{thakur2021beir}. Following the settings of \citet{gao-etal-2021-simcse}, we use Spearman's correlation to measure the performance of different approaches. We choose BERT$_{base}$~\cite{devlin-etal-2019-bert} and RoBERTa$_{base}$~\cite{Liu2019RoBERTaAR} as our backbone models $G_\phi$. 
For the unlabeled sentences, we use the premises of NLI from \citet{gao-etal-2021-simcse} as unlabeled sentences, and ensure the data volume used by MultiCSR is equivalent to that of SimCSE. For the main results in Table \ref{tab:unsup_ours}, we include only the results with NLI premises following ~\citet{zhang-etal-2023-contrastive-learning}. In addition, we also discuss the impact of different data resources in Section \ref{sec:resource}. While our framework is general and could be combined with more advanced algorithms as well, we utilize SimCSE as our main backbone. In Appendix \ref{appendix_backbone}, we include more experimental results of applying MultiCSR to different backbones.

We utilize the corresponding versions of SimCSE (i.e., unsupervised and supervised, BERT$_{base}$ and RoBERTa$_{base}$) as $P_\eta$ in different settings. For LLM $C_\theta$, we include the results of Flan-T5-XL(3B) to show that, with MultiCSR, a relatively tinier and less advanced LLM can even outperform ChatGPT, while applying to ChatGPT achieves a better state-of-the-art performance. In the experiments with ChatGPT, since their logits can not be acquired, we utilize only the last two stages of MultiCSR. We also discuss the effect of smoothing coefficient $\omega$, self-curation thresholds $\alpha$, $\beta$ and $\gamma$ in Appendix \ref{appendix_omega} and \ref{appendix_filter} separately, and only include the results with $\omega=0.3$, $\alpha=3$, $\beta=3$ and $\gamma=1$ in main results. We fine-tune our model with a batch size of 256, $\tau=0.05$ and $\sigma=0.9$, and choose the best model parameters $\phi$ based on the development performance from STS-Benchmark following \citet{gao-etal-2021-simcse}. We conduct ablation studies on $\sigma$ in Appendix \ref{appendix_threshold}. We include the performance on transfer tasks and BEIR tasks in Appendix \ref{appendix_transfer}. We also evaluate MultiCSR in a supervised setting by directly combining our generated corpus with labeled NLI corpus. Because of our proposed contrastive in-batch training, the \textit{false-negative} issue will not be raised with this direct combination. We include a more detailed analysis and our supervised performance in Appendix \ref{appendix_demons}.

\begin{table}[t]
\centering
\renewcommand\arraystretch{1.1}
\setlength{\tabcolsep}{3.5pt}
\begin{tabular}{lcc}
\hline
\textbf{Method} & \textbf{Spearman's}& \textbf{$\Delta$} \\
\hline
\hline
MultiCSR$_{\rm RoBERTa}$& \textbf{85.82} & 0.00 \\
\ \ w/o \textbf{\textcolor{stage1}{stage 1}} & 84.06 & -1.76\\
\ \ w/o \textbf{\textcolor{stage2}{stage 2}} & 83.98 & -1.84\\
\ \ w/o \textbf{\textcolor{stage3}{stage 3}} & 84.79 & -1.03\\
\ \ w/o \textbf{\textcolor{stage2}{stage 2}} \& \textbf{\textcolor{stage3}{3}} & 82.74 & -3.08\\
\ \ w/o \textbf{\textcolor{stage1}{stage 1}} \& \textbf{\textcolor{stage2}{2}} & 79.86 & -5.96\\
\ \ w/o \textbf{\textcolor{stage1}{stage 1}} \& \textbf{\textcolor{stage3}{3}} & 82.45 & -3.37\\
\ \ w/o \textbf{\textcolor{stage1}{stage 1}} \& \textbf{\textcolor{stage2}{2}} \& \textbf{\textcolor{stage3}{3}} & 75.51 & -10.31\\
\hline
\end{tabular}
\vspace{-1mm}
\caption{Ablation studies of different components in MultiCSR (Flan-T5) based on SimCSE$_{\rm RoBERTa}$. The results are based on the development set of STS-B. MultiCSR w/o \textbf{\textcolor{stage1}{stage 1}}\&\textbf{\textcolor{stage2}{2}}\&\textbf{\textcolor{stage3}{3}}: training SimCSE with the raw generation of Flan-T5-XL.} \label{tab:ablation}
\vspace{-2mm}
\end{table}

\paragraph{Baselines}
We compare our method with many strong baselines including ConSERT~\cite{yan-etal-2021-consert}, SimCSE~\cite{gao-etal-2021-simcse}, DiffCSE~\cite{chuang-etal-2022-diffcse}, PromptBERT~\cite{jiang-etal-2022-promptbert}, InfoCSE~\cite{wu-etal-2022-infocse}, RankEncoder~\cite{seonwoo-etal-2023-ranking}, RankCSE~\cite{liu-etal-2023-rankcse} and a post-processing method BERT-whitening~\cite{Su2021WhiteningSR}. To make comparison between different corpus construction methods, we further compare MultiCSR with SynCSE~\cite{zhang-etal-2023-contrastive-learning}, which directly leverages the generated sentences of ChatGPT to train SimCSE.

\subsection{Main Results}
From the results shown in Table \ref{tab:unsup_ours}, we have the following observations: (1) From the comparison between our approach and previous strong baselines, MultiCSR can significantly enhance the base model SimCSE and raises the averaged Spearman's correlation from 76.25\% and 76.57\% to 80.93\% and 81.62\% respectively, achieving a better state-of-the-art performance. It is also important to note that MultiCSR is general and data-oriented, and can be easily applied to various base models and improves their performance as shown in Appendix \ref{appendix_backbone}. Although we build our method over SimCSE, MultiCSR can achieve comparable or even better results than the strong and competitive models such as RankCSE, which demonstrates the effectiveness of our method.
(2) When comparing with SynCSE, which also leverages LLMs for enhancing sentence representation learning, MultiCSR shows to be more effective. By utilizing only Flan-T5-XL(3B), our approach can even outperform SynCSE with ChatGPT on RoBERTa$_{base}$, which also provides an opportunity for broader open-source but less advanced LLMs, rather than blindly pursuing larger and better LLMs (which are mostly closed-source). In addition, MultiCSR can still be valid when applied to ChatGPT, which demonstrates our claim that a curated training corpus is necessary for effectively training a contrastive learning method.

\section{Analysis} \label{sec:analysis}
\subsection{Ablation Studies}
We investigate the impact of each component in MultiCSR and report the performance in Table \ref{tab:ablation}. The results of MultiCSR w/o {stage 1}\&{2}\&{3} demonstrate our motivation that the raw generated content from LLMs may not satisfy the requirement of training a contrastive learning method. The comparison between this result and those of removing two stages shows that each component plays a crucial role in enhancing the base model.

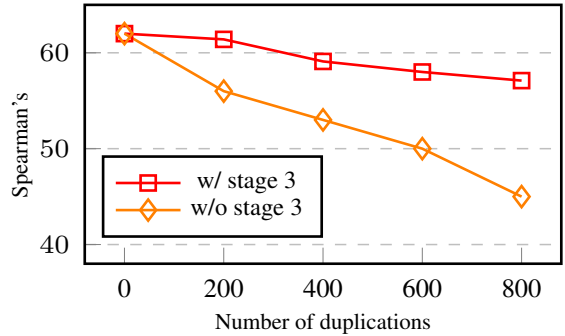
\begin{figure}
\centering
\resizebox{1\linewidth}{!}{
\begin{tikzpicture}
\pgfplotsset{width = 6cm, height = 4cm}
    \begin{axis}[
        xlabel={\tiny Number of duplications},
        ylabel={\tiny Spearman's},
        line width=0.7pt,
        ymax=65,
        ymin=38,
        y label style={yshift=-0.65cm, xshift=-0.1cm},
        x label style={yshift=0.2cm, xshift=0.0cm},
        label style={font=\fontsize{7}{1}\selectfont},
        xtick = {1,2,3,4,5},
        xticklabels = {0, 200, 400, 600, 800},
        xticklabel style = {font=\fontsize{7}{1}\selectfont, rotate=0,},
        yticklabel style = {font=\fontsize{7}{1}\selectfont},
        xtick pos = left,
        ytick pos = left,
        legend pos = south east,
        legend style={font=\fontsize{6.5}{1}\selectfont, row sep=-0.1cm,/tikz/every odd column/.append style={column sep=0.01cm}},
        legend style={at={(0.5,0.1)},legend columns=1}, 
        ymajorgrids = true,
        grid style=dashed,
    ]
    \addplot [mark=square, mark size=2pt, color=red] plot coordinates {
    (1, 62) (2, 61.4) (3, 59.1) (4, 58.0) (5, 57.1)};
    \addlegendentry{w/ stage 3};
    \addplot [mark=diamond, mark size=2.8pt, color=orange] plot coordinates {
    (1, 62) (2, 56) (3, 53) (4, 50) (5, 45)};
    \addlegendentry{w/o stage 3};
    \end{axis}
\end{tikzpicture}
}
\vspace{-7mm}
\caption{Performance comparison between with and without our constrastive in-batch training stage, with the number of duplications and Spearman's scores on the development set of STS-B reported. 
}
\vspace{-3mm}
\label{fig:stage3}
\end{figure}

Among them, since stage 1 and stage 2 are strictly controlling the quality of each generation or each sentence pair, they seem to be more important than stage 3. However, they are in charge of different stages of training a contrastive learning method like SimCSE. In a training batch with $N$ sentence pairs, during the data generation (stage 1) and sentence pair construction (stage 2), they only check whether $2N$ sentence pairs are qualified. But for stage 3, it will consider whether $2N(N-1)$ sentence pairs involve \textit{false negatives} or not. Moreover, the control of stage 3 will not be too strict for a high-quality dataset like NLI where most premises are not similar to each other as shown in the statistical results in Appendix \ref{appendix_demons}. For a dataset where only limited data are available, whether using stage 3 or not will result in a huge performance gap. To demonstrate this, we sample a sentence pair from $10K$ ground truth sentence pairs of NLI, and duplicate it for certain times and combine this set with other pairs together to train a SimCSE model. The results are shown in Figure \ref{fig:stage3}. It is important to note that all the sentence pairs are in good condition and will survive after stage 1\&2. In this scenario, only our proposed contrastive in-batch training stage can be helpful. In our supervised setting, by utilizing this in-batch masking, we can directly combine the generated corpus with the ground truth NLI dataset as the training corpus, without suffering from the \textit{false-negative} issue, as shown in Appendix \ref{appendix_demons}.

\begin{table}[t]
\centering
\renewcommand\arraystretch{1.1}
\setlength{\tabcolsep}{6pt}
\begin{tabular}{llc}
\hline
\textbf{Method} & \textbf{Resource} & \textbf{Spearman's}\\
\hline
\hline
\multicolumn{3}{c}{\textit{BERT-base}} \\
\hline
\multirow{2}{*}{SimCSE} & NLI* & 76.7\\
& Wikipedia & 81.9 \\
\hline
\rowcolor{shadecolor}  & NLI & \textbf{84.6} \\
\rowcolor{shadecolor} \multirow{-2}{*}{MultiCSR} & Wikipedia & 83.5 \\

\hline 
\multicolumn{3}{c}{\textit{RoBERTa-base}} \\
\hline
\multirow{2}{*}{SimCSE} & NLI* & 81.1\\
& Wikipedia & 82.9 \\
\hline
\rowcolor{shadecolor}  & NLI & \textbf{85.8} \\
\rowcolor{shadecolor} \multirow{-2}{*}{MultiCSR} & Wikipedia & 85.2 \\
\hline 
\end{tabular}
\vspace{-1mm}
\caption{Performance comparison of using different data resources on the development set of STS-B. *: since we only utilize NLI premises rather than (premise, entailment, contradiction) triplets, we use only these premises to train the unsupervised version of SimCSE.} \label{tab:data_resource}
\vspace{-3mm}
\end{table}

Through the analysis of our ablation studies, each component is necessary for MultiCSR to ensure that a contrastive learning method is only trained on a high-quality and effective corpus for better sentence representation learning.

\subsection{The Impact of Different Data Resources} \label{sec:resource}
In Section \ref{sec:experiment_setup}, we have introduced that we utilize the NLI premises as the unlabeled sentences, following previous work. As a high-quality and crowd-sourced dataset, sentences from NLI datasets will have less lexical overlap and be in good conditions. It is also important to show that our method can still be valid for sentences from different domains, which are usually noisier but more accessible. Thus, we further utilize the same $10^6$ randomly sampled sentences of Wikipedia from \citet{gao-etal-2021-simcse} as the unlabeled sentences. We follow the same settings of main results in Table \ref{tab:unsup_ours}, and the results are shown in Table \ref{tab:data_resource}. From the results of SimCSE with different resources, we can see that training a contrastive sentence embedding model will place a requirement on the data volume, since the number of sentences for NLI is 0.28M and much smaller than 1.00M of Wikipedia samples. Thus, as a flexible corpus construction framework, MultiCSR can largely outperform the models trained with only these unlabeled sentences, which demonstrates our effectiveness and robustness across different data resources. 

When we compare the results of MultiCSR with different data resources, we can see that models trained with NLI premises essentially outperform those with Wikipedia sentences. We assume this may due to the fact that, NLI premises are typically short statements that express the relationship between two entities or concepts, while Wikipedia sentences are usually longer and provide more detailed information about a particular topic and are more factual which requires a broader range of language understanding abilities to write their entailment or contradiction hypotheses. There might require a more appropriate way to generate the positives and negatives of Wikipedia sentences, which we leave as future work.

\section{Discussion: Different Strategies for Self-Curation and In-Batch Training}

In Section \ref{sec:Methodology}, we have introduced in detail how we perform contrastive sentence pair construction with self-curation (denoted as stage 2) and contrastive in-batch training (as stage 3), and both of them are completed by measuring the similarities. It seems that we can use LLMs' and also pre-trained models' similarities for both stages. However, we have our intuition behind these designs.

Our design was firstly motivated by efficiency considerations. Before introducing the selection for stage 2, we will explain why we choose to utilize pre-trained models in stage 3. There are multiple implementations for this stage. For example, we can also adopt a method similar to stage 2, having the LLM compute similarity scores and provide masks. This allows the entire framework to rely only on the LLMs for data quality control. However, due to the slow inference speed of the LLM, we must calculate and store the similarity scores for each potential pair in advance with LLMs, to not impact the training speed of the base model. In addition, since in-batch training data are randomly sampled, all sentence pairs should be considered. Given a training corpus with $M$ triplets and a batch size of $N$, the number of similarity scores that LLMs must generate in advance is $2M(M-1)$, but only $2M(N-1)$ of them will be used in stage 3. Since $M$ is usually 3 orders of magnitude larger than $N$, a significant amount of computational resources would be wasted. Therefore, we choose to use a pre-trained model to generate masks dynamically in the stage 3 which will have little impact on the training speed. We also provide some cost analysis for stage 3 in Appendix \ref{appendix_cost_analysis}. 

Furthermore, if we try to utilize the similar strategy of stage 3 in stage 2, we would use a pre-trained sentence embedding model corresponding to the base model to generate similarity scores. This means that when we want to train or evaluate on a different backbone, we would need to re-run the same self-curation process on the data generated in stage 1 again. This also leads to a waste of computational resources. Besides, using an LLM to evaluate a sentence triplet could be completed simultaneously with stage 1. When we choose to use LLMs in stage 2 and a pre-trained model in stage 3, stage 1\&2 and stage 3 will be responsible for the dataset construction process and during-training process, respectively. The high-quality training corpus produced after stage 1\&2 can be used in training or evaluation of various base models in similar settings in stage 3. We further include some related case studies in Appendix \ref{appendix_example}.

Moreover, using a pre-trained sentence embedding model in stage 2 will contradict our framework's objective in some scenarios. As previously mentioned, we would choose a pre-trained model corresponding to different base models to compute the similarity scores. When the checkpoints of a certain pre-trained model have not been released, we can even train such models using the high-quality training corpus, with the obtained model used to provide masks in stage 3. However, this behavior assumes that we already have a high-quality training corpus, which would be a significant challenge with only the corpus generated from stage 1, as demonstrated in the results of our ablation study's MultiCSR w/o stage 2\&3. And stage 2 is precisely designed as a necessary stage to improve the quality of training corpus. Hence, attempting to introduce a pre-trained model in stage 2 would require either a corresponding released checkpoint or high-quality training data. In more extreme scenarios, such as the low-resource scenarios, there is no pre-trained sentence embedding model available, and the sentences generated with only contrastive generation strategy is not sufficient to form a high-quality training corpus. 

To demonstrate this, we further apply our MultiCSR to low-resource languages. We conduct experiments on a rather small language: Tagalog (TL). For training corpus, we leverage TED2020 from \citet{reimers-gurevych-2020-making} with $1167$ translated sentence-pairs, and use only the sentences in TL. The evaluation is performed by finding the most similar sentence inside a corpus using cosine similarity, with $1000$ test pairs from LASER~\cite{artetxe-schwenk-2019-massively}. The results are shown in Table \ref{tab:low_source}. For a fair comparison, the only difference of ChatGPT and MultiCSR is whether utilizing our proposed self-curation during contrastive sentence pair construction. From the results, we can observe significant improvement from utilizing methods with LLMs, and the performance can be further enhanced with MultiCSR. The evaluation of applying MultiCSR to this challenging scenario demonstrates the flexibility of our framework, enabling its broad applications across different domains and even languages. These experiments also demonstrate that, the conditions for introducing a pre-trained sentence embedding model in stage 2 cannot always be guaranteed. Therefore, we suggest to consider the proposed strategies in a way we present in our paper.

\begin{table}[t]
\centering
\renewcommand\arraystretch{1.1}
\setlength{\tabcolsep}{8pt}
\begin{tabular}{lc}
\hline
\textbf{Method} & \textbf{Accuracy\%}\\
\hline
\hline
SimCSE$_{\rm BERT}$  & 72.6 \\
\hline
+ \textbf{\textcolor{stage1}{stage 1}}  & 77.9 \\
\rowcolor{shadecolor}+ \textbf{\textcolor{stage1}{stage 1}} \& \textbf{\textcolor{stage2}{2}}  & 84.2 \\
\hline 
\end{tabular}
\vspace{-1mm}
\caption{Performance comparison of different methods on the low-resource language Tagalog. For SimCSE, we only use the unlabeled sentences to train its unsupervised version. We utilize ChatGPT for data generation.} \label{tab:low_source}
\vspace{-2mm}
\end{table}

Apart from these concerns, when only taking the effectiveness into consideration, the similarity scores provided directly by the LLMs are not as accurate as those from a pre-trained sentence embedding model, as shown in Section \ref{sec:related_work_llm} and Table \ref{tab:llm_scoring}. This is also our motivation for why we choose not to let the base model merely imitate the scores computed by LLMs. In addition, there are some alternatives for self-curation and in some scenarios, they can even be combined together or iteratively utilized in the same stage. We believe that our self-curation strategy in stage 2 is a practical and thoughtful solution, and we are also committed to continuously explore other potential alternatives.

\section{Conclusion}
In this paper, we introduce MultiCSR, a novel framework to contrastively refine the generation of LLMs at distinct stages of training a contrastive learning method, ensuring only high-quality and suitable sentence pairs are utilized during the model training. Experiments on standard STS tasks and several downstream tasks demonstrate the effectiveness of MultiCSR. Extensive analysis shows the potential of our work and we hope to inspire future work in achieving better sentence representation learning with LLMs.

\section*{Acknowledgements} 
This work was substantially supported by DAMO Academy through DAMO Academy Research Intern Program, and is partially supported by the grant RS-INSUR-00027-E0901-S00.

\section*{Limitations} \label{sec:limitation}
Despite the effectiveness of MultiCSR, there are still some potential directions worth exploring and we leave as future work. Firstly, as introduced in \ref{sec:filter}, we employ pre-defined rules to control the absolute and relative similarities of each pair $(\boldsymbol{x}_+, \boldsymbol{x}_-)$ during the contrastive sentence pair construction. However, a composite model can be utilized here. As our generated sentence pairs are in the same formats of NLI datasets~\cite{zhang-etal-2023-contrastive-learning}, we can actually use a NLI classifier or its combination with LLMs for self-curation. Nevertheless, utilizing a pre-trained NLI classifier also poses a requirement for ground truth NLI corpus, contradicting the main purpose of performing unsupervised learning. Thus, we leave this as future work and hope to propose some alternative self-curation strategies. Secondly, in the first stage of our framework, we perform contrastive generation with logits acquired from different instructions. However, for closed-source LLMs, acquiring their logits is impractical. Thus, a prompting methods that can be incorporated into refining the generation of LLMs will be valuable. Recently, a contemporary method, self-correction, has been proposed to address this issue. However, the effectiveness of the techniques in this kind often depends on the fortuitous alignment of prompts or initial conditions, making them labor-intensive, highlighting more efficient approaches. To sum up, these limitations also illustrate the great potential of our method. It is expected to be applied to various domains and better serve more downstream tasks.

\bibliography{anthology,custom}

\appendix

\clearpage

\section*{Appendix} \label{sec:appendix}

\section{Instruction Prompts} \label{appendix_prompt}
In this section, we give the details of our instruction prompts used in $C_\theta$ text generation and self-curation. In addition, for fair comparison with SynCSE~\cite{zhang-etal-2023-contrastive-learning}, we utilize the same entailment and contradiction prompts for generation. During the first stage of contrastive generation, we will randomly select one from the entailment prompts and the other one from the contradiction prompts, and acquires the next-token logits with the logits derived from these two instructions.

\begin{tcolorbox}[title=Entailment Prompt 1]
\small
Please paraphrase the input sentence or phrase, providing an alternative expression with the same meaning.
\end{tcolorbox}

\begin{tcolorbox}[title=Entailment Prompt 2]
\small
Rewrite the following sentence or phrase using different words and sentence structure while preserving its original meaning.
\end{tcolorbox}

\begin{tcolorbox}[title=Entailment Prompt 3]
\small
Create a sentence or phrase that is also true, assuming the provided input sentence or phrase is true.
\end{tcolorbox}

\begin{tcolorbox}[title=Entailment Prompt 4]
\small
Please provide a concise paraphrase of the input sentence or phrase, maintaining the core meaning while altering the words and sentence structure. Feel free to omit some of the non-essential details like adjectives or adverbs.
\end{tcolorbox}

\begin{tcolorbox}[title=Contradiction Prompt 1]
\small
Revise the provided sentence by swapping, changing, or contradicting some details in order to express a different meaning, while maintaining the general context and structure.
\end{tcolorbox}

\begin{tcolorbox}[title=Contradiction Prompt 2]
\small
Generate a slightly modified version of the provided sentence to express an opposing or alternate meaning by changing one or two specific elements, while maintaining the overall context and sentence structure.
\end{tcolorbox}

\begin{tcolorbox}[title=Contradiction Prompt 3]
\small
Transform the input sentence by adjusting, altering, or contradicting its original meaning to create a logical and sensible output sentence with a different meaning from the input sentence.
\end{tcolorbox}

\begin{tcolorbox}[title=Contradiction Prompt 4]
\small
Generate a sentence that conveys a altering, contrasting or opposite idea to the given input sentence, while ensuring the new sentence is logical, realistic, and grounded in common sense.
\end{tcolorbox}

While we utilize an LLM to measure the semantic similarity, we will use the prompt as the following:
\begin{tcolorbox}[title=Measuring Similarity Prompt]
\small
Scoring the semantic similarity of the following sentences between 0.0 and 5.0, 5.0 means they have the same meaning, 0.0 means they are completely different: (a) ``$\{\boldsymbol{x}\}$'', (b) ``$\{\boldsymbol{x}_y\}$'':
\end{tcolorbox}

\section{MultiCSR with Various Backbones} \label{appendix_backbone}
In our main results from Table \ref{tab:unsup_ours}, we improve the performance of SimCSE by a large margin. It is important to note that MultiCSR is general and can be uniformly applied to different backbones. Thus, in this section, we also conduct experiments with representative contrastive sentence representation learning methods PromptBERT~\cite{jiang-etal-2022-promptbert}, InfoCSE~\cite{wu-etal-2022-infocse}, RankEncoder~\cite{seonwoo-etal-2023-ranking} and RankCSE~\cite{liu-etal-2023-rankcse}. The results shown in Table \ref{tab:gpt} show that, our approach can consistently improve their performance by a large margin, which also demonstrates the strong generalization ability of our MultiCSR.

\begin{table}[t]
\centering
\renewcommand\arraystretch{1.1}
\setlength{\tabcolsep}{20pt}
\vspace{-2mm}
\begin{tabular}{lc}
\hline
\textbf{Model} & \textbf{Avg.}\\
\hline
\hline
\multicolumn{2}{c}{\textit{BERT-base}} \\
\hline
\hline
SimCSE& 76.25 \\
+ MultiCSR&80.02\\
\hline
PromptBERT& 78.54 \\
+ MultiCSR&80.33\\
\hline
InfoCSE& 78.85 \\
+ MultiCSR&80.45\\
\hline
RankEncoder& 80.07 \\
+ MultiCSR&80.52\\
\hline
RankCSE& 80.36 \\
+ MultiCSR&80.79\\
\hline 
\hline

\multicolumn{2}{c}{\textit{RoBERTa-base}} \\
\hline 
\hline
SimCSE& 76.57 \\
+ MultiCSR&81.01\\
\hline
PromptRoBERTa& 79.15 \\
+ MultiCSR&81.24\\
\hline
RankCSE&79.81 \\
+ MultiCSR&81.45\\
\hline
\end{tabular}
\caption{Performance comparison on different backbone contrastive sentence representation learning methods between with and without MultiCSR.
} \label{tab:gpt}
\end{table}

\section{Effect of Smoothing Coefficient $\omega$} \label{appendix_omega}
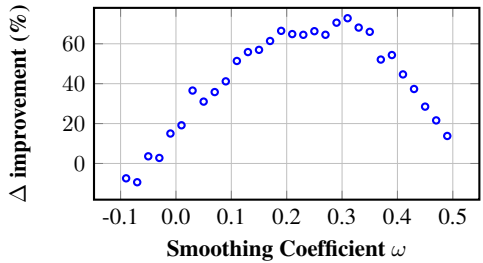
\begin{figure}[!t]
  \centering
  {
      \begin{tikzpicture}[scale=0.79]

      \begin{axis}[
          line width=1pt,
          height=0.3\textwidth,
          width=0.50\textwidth,
          xlabel = \textbf{Smoothing Coefficient $\omega$},
          ylabel = \textbf{$\Delta$ improvement (\%)},
          xtick={-0.1, 0.0, 0.1, 0.2, 0.3, 0.4, 0.5},
          xticklabels={-0.1, 0.0, 0.1, 0.2, 0.3, 0.4, 0.5},
          ymax=78,
          grid=major,
          legend pos=south east,
      ]

      \addplot+[only marks,color=blue,mark=o, mark size=1.5pt] coordinates {
( -0.09  ,  -7.449516917728406 ) ( -0.07  ,  -9.40632156342002 ) ( -0.05  ,  3.582614263088567 ) ( -0.03  ,  2.77016147306676 ) ( -0.01  ,  15.011635268340505 ) ( 0.01  ,  19.14166542702913 ) ( 0.03  ,  36.54873957640397 ) ( 0.05  ,  31.0071527523026 ) ( 0.07  ,  35.78800402549445 ) ( 0.09  ,  41.16434152245043 ) ( 0.11  ,  51.38099868558881 ) ( 0.13  ,  55.8436378795229 ) ( 0.15  ,  56.94673276803748 ) ( 0.17  ,  61.40053699754996 ) ( 0.19  ,  66.47624105365612 ) ( 0.21  ,  64.85814158794021 ) ( 0.23  ,  64.49326605045303 ) ( 0.25  ,  66.3083024716849 ) ( 0.27  ,  64.513799784782 ) ( 0.29  ,  70.57318412955459 ) ( 0.31  ,  72.809597179983335 ) ( 0.33  ,  68.11367539911883 ) ( 0.35  ,  65.99277711540144 ) ( 0.37  ,  52.08675029424459 ) ( 0.39  ,  54.34904608037005 ) ( 0.41  ,  44.64612748076706 ) ( 0.43  ,  37.32956230834168 ) ( 0.45  ,  28.493107870197658 ) ( 0.47  ,  21.589472342807166 ) ( 0.49  ,  13.78332675276033 )
      };
      \end{axis}

      \end{tikzpicture}
      }
  \vspace{-3mm}
  \caption{Generation improvement by incorporating noisy instruction with $l_t-\omega*\hat{l}_t$ into our contrastive generation stage.} \label{fig:omega}
  \vspace{-5.5mm}
\end{figure}

Figure \ref{fig:omega} shows the influence of the hyperparameter $\omega$ on our method’s performance. This parameter adjusts the smoothness of logits derived from noisy instructions. For all the evaluation, we utilize opposite prompts as the noisy instructions. We sample $100$ sentences from the NLI premises and conduct human evaluation. From the results, we can see that performance tends to decline with negative $\omega$ values, as the model becomes increasingly biased toward the noisy instruction. Conversely, excessively positive values (above 0.4) lead to a quick deterioration in performance. Interestingly, the model’s performance stabilizes between 0.2 and around 0.4, indicating a certain level of robustness to variations in $\omega$ within this range. In our main results, we consistently utilize $\omega$ as $0.3$.

\begin{table}[!ht]
\centering
\small
\renewcommand\arraystretch{1.1}
\setlength{\tabcolsep}{12pt}
\vspace{-2mm}
\begin{tabular}{llllc}
\hline
\textbf{$\alpha$}&\textbf{$\beta$}&\textbf{$\gamma$} &\textbf{$|\mathcal{T}|$}&\textbf{Spearman’s}\\
\hline
\hline
5 & 0 & - & 30.7k & 82.42 \\
5 & 1 & - & 37.7k & 82.03 \\
5 & 2 & - & 37.9k &  81.67\\
5 & 3 & - & 47.8k &  82.69\\
5 & 4 & - & 108.9k &  79.77\\
\hline
4 & 0 & - & 48.8k &  84.96\\
4 & 1 & - & 60.4k &  84.80\\
4 & 2 & - & 60.7k &  85.17\\
4 & 3 & - & 72.9k &  85.69\\
\hline
3 & 0 & - & 51.6k &  84.31\\
3 & 1 & - & 63.7k &  84.48\\
3 & 2 & - & 64.1k &  85.12\\
\hline
2 & 0 & - & 51.6k &  84.29\\
2 & 1 & - & 63.8k &  84.72\\
\hline
1 & 0 & - & 54.4k & 83.13 \\
\hline
\hline
4 & 4 & 1 &134.0k &  82.92 \\
4 & 4 & 2 &70.6k & 83.30\\
4 & 4 & 3 &60.6k & 85.00\\
4 & 4 & 4 &55.8k & 85.19\\
\hline
3 & 3 & 1 &76.4k & \textbf{85.82}\\
3 & 3 & 2 & 73.9k& 84.70\\
3 & 3 & 3 & 63.4k& 85.49\\
3 & 1 & 3 & 63.2k& 84.35\\
\hline
2 & 1 & 2 & 63.8k& 84.90\\
2 & 2 & 1 & 64.1k& 85.13\\
2 & 2 & 2 & 64.0k& 85.36\\
\hline
1 & 1 & 1 & 66.6k& 85.42\\
\hline
\end{tabular}
\caption{Studies of different self-curation strategies in MultiCSR. The results are the development performance of STS-B.
} \label{tab:filters}
\vspace{-5mm}
\end{table}

\section{Self-Curation Strategies} \label{appendix_filter}
In Section \ref{sec:filter}, we briefly introduce how we select ``high-quality'' triplets like $(\boldsymbol{x}, \boldsymbol{x}_+, \boldsymbol{x}_-)$ after we get the semantic similarity scores from the LLM $C_\theta$ as $a$ and $b$:
\begin{equation}
(\boldsymbol{x}, \boldsymbol{x}_+, \boldsymbol{x}_-) \in \mathcal{T},\, {\rm if} \begin{cases}
a \geq \alpha\\
b \leq \beta\\
a \geq b + \gamma
\end{cases}.
\end{equation}
We firstly conduct extensive experiments on different combinations of hyperparameters $\alpha$, $\beta$ and $\gamma$. The results are shown in Table \ref{tab:filters}. Based on the results from Table \ref{tab:llm_scoring}, we know that directly utilizing LLMs to measure the similarities may not be a wise choice, but their provided signals can still help us improve the quality of generated corpus. From the results in Table \ref{tab:filters}, we can see that choices that are not too extreme (e.g., set $\alpha$ to 5 or $\beta$ to 0) tend to give better results, which also demonstrate our concern that ``\textit{hard negative}'' and ``\textit{hard positive}'' are always important to provide sufficient challenges to learn a better contrastive learning model.

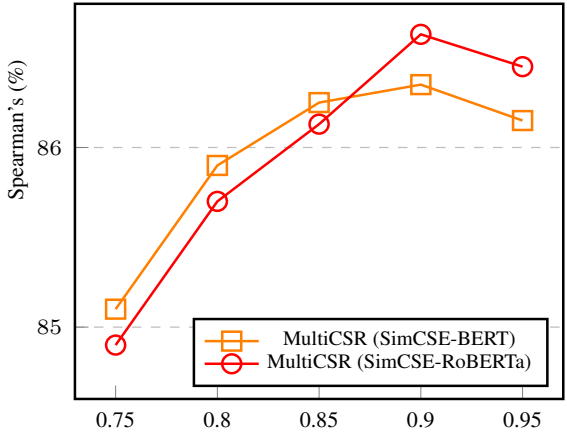
\begin{figure}
\centering
\resizebox{1\linewidth}{!}{
\begin{tikzpicture}[font=\Large]
\pgfplotsset{width = 7cm, height = 6cm}
    \begin{axis}[
        ylabel={Spearman's (\%)},
        line width=0.8pt,
        ymax=86.8,
        ymin=84.6,
        y label style={yshift=-0.6cm, xshift=0.8cm},
        x label style={yshift=0.4cm, xshift=2.8cm},
        label style={font=\fontsize{7}{1}\selectfont},
        xtick = {0.7,0.75,0.8,0.85,0.9,0.95,1},
        ytick = {85, 86, 87},
        xticklabels = {0.7,0.75,0.8,0.85,0.9,0.95,1},
        xticklabel style = {font=\fontsize{7}{1}\selectfont, rotate=0,},
        yticklabel style = {font=\fontsize{7}{1}\selectfont},
        xtick pos = left,
        ytick pos = left,
        legend pos = south east,
        legend style={font=\fontsize{6.5}{1}\selectfont, row sep=-0.1cm,/tikz/every odd column/.append style={column sep=0.01cm}},
        ymajorgrids = true,
        grid style=dashed,
    ]
    \addplot [mark=square, mark size=3pt, color=orange] plot coordinates {
    (0.75, 85.1) (0.80, 85.9) (0.85, 86.25) (0.9, 86.35) (0.95, 86.15)};
    \addlegendentry{MultiCSR (SimCSE-BERT)};
    \addplot [mark=o,  mark size=3pt, color= red] plot coordinates {
    (0.75, 84.9) (0.80, 85.7) (0.85, 86.13) (0.9, 86.63) (0.95, 86.45)};
    \addlegendentry{MultiCSR (SimCSE-RoBERTa)};
    \end{axis}
\end{tikzpicture}
}
\vspace{-2mm}
\caption{The performance of MultiCSR with different mask indicator thresholds. The results are the development performance of STS-B.}
\vspace{-3mm}
\label{fig:threshold}
\end{figure}

\section{Ablation Study on Mask Indicator Threshold} \label{appendix_threshold}
In this section, we perform an ablation study of setting different mask indicator thresholds used in Section \ref{sec:loss} to mask in-batch false negatives. The results are presented in Figure \ref{fig:threshold}. As shown in the figure, introducing the mask indicator mechanism significantly improves the results. When the threshold is set too low, the performance drops as expected, due to the fact that too many sentences are masked and few challenging inputs are given to the model. Meanwhile, the purpose of filtering out false negative sentences is not fulfilled if the threshold chosen is too high. Based on the experimental results on STS-B development set, we set the mask indicator threshold to be 0.9 across our main context.

\section{Performance on Transfer Tasks and BEIR Tasks} \label{appendix_transfer}

We also evaluate our approach on the following transfer tasks: MR~\cite{pang2005seeing_mr}, CR~\cite{hu2004mining_cr}, SUBJ~\cite{pang2004sentimental_subj}, MPQA~\cite{wiebe2005annotating_mpqa}, SST-2~\cite{socher2013recursive_sst-2}, TREC~\cite{voorhees2000building_trec} and MRPC~\cite{dolan-brockett-2005-automatically-mrpc}. In these tasks, a classifier is trained on top of sentence representations produced by different methods. We reported the detailed performance of MultiCSR in Table \ref{tab:transfer}. From the experimental results, we can see that MultiCSR achieves the best performance on RoBERTa$_{base}$ and comparable results on BERT$_{base}$, which also shows the potential of our method to be applied to different downstream tasks.

\begin{table}[!t]
\centering
\small
\renewcommand\arraystretch{1.1}
\setlength{\tabcolsep}{9.5pt}
\begin{tabular}{lcc}
\hline
\textbf{Dataset/Model}&\textbf{SimCSE} &\textbf{MultiCSR}\\
\hline
\hline
MS MARCO & 0.230 & 0.270\\
TREC-COVID & 0.298 & 0.352\\
NFCorpus & 0.125 & 0.149\\
NQ & 0.141 & 0.182\\
HotpotQA & 0.219 & 0.242\\
FiQA-2018 & 0.091 & 0.123\\
ArguAna & 	0.386  & 	0.417\\
Touché-2020 & 	0.089	 & 0.129\\
CQADupStack & 	0.155 & 	0.191\\
Quora & 	0.768 & 	0.788\\
DBPedia & 	0.191 & 	0.204\\
SCIDOCS & 	0.075 & 	0.097\\
FEVER & 	0.122 & 	0.174\\
Climate-FEVER & 	0.111 & 	0.140\\
SciFact & 	0.329	 & 0.381\\
\hline
\textbf{Avg.} & 	0.222 (1x) & 	\textbf{0.256 (1.15x)}\\
\hline
\end{tabular}
\caption{Performance comparison on zero-shot information retrieval tasks BEIR.
} \label{tab:beir}
\vspace{-5mm}
\end{table}

To further evaluate our framework on real-world applications, we also test the performance of MultiCSR on zero-shot information retrieval tasks BEIR~\cite{thakur2021beir} and include the results in Table \ref{tab:beir}. We directly utilize our trained checkpoint to test and no in-domain data is used. The results are shown in the following table with nDCG\@10 scores reported, following BEIR. We use all the tasks public in BEIR. For CQADupStack, we evaluate each StackExchange subforum separately and report the overall average scores. The results show that our method can consistently improve the performance of SimCSE on these tasks, even before we use any unlabeled in-domain data. We believe that better semantic similarity performance is beneficial to the measurement of sentence-pair relationship, and our method can help the base model capture richer semantic information.

\begin{table*}[!ht]
\centering
\renewcommand\arraystretch{1.1}
\small
\setlength{\tabcolsep}{8pt}

\begin{tabular}{lcccccccc}
\hline
\textbf{Model} & \textbf{MR} & \textbf{CR} & \textbf{SUBJ} & \textbf{MPQA} & \textbf{SST-2} & \textbf{TREC} & \textbf{MRPC} & \textbf{Avg.}\\
\hline
\hline
\multicolumn{9}{c}{\textit{BERT-base}} \\
\hline
SimCSE &81.18& 86.46& 94.45& 88.88& 85.50& \textbf{89.80}& 74.43& 85.81 \\
DiffCSE* & 81.76& 	86.20& 	94.76& 	89.21& 	86.00& 	87.60& 	75.54& 	85.87 \\
PromptBERT & 80.74& 85.49& 93.65& 89.32& 84.95& 88.20& 76.06& 85.49\\
RankCSE& \textbf{83.07}  & \textbf{88.27}  & \textbf{95.06} &  89.90  & 87.70  & 89.40 &  \textbf{76.23}  & \textbf{87.09}\\

\hline

MultiCSR (SimCSE)& 82.70 & 88.15 & 94.97 & \textbf{90.08} & 86.87 & 87.70 & 75.46 & 86.56\\

\hline

\multicolumn{9}{c}{\textit{RoBERTa-base}} \\
\hline
SimCSE &81.04& 87.74& 93.28& 86.94& 86.60& 84.60& 73.68& 84.84 \\
DiffCSE* & 82.42	& 88.34& 	93.51& 	87.28& 	87.70& 	86.60& 	76.35& 	86.03\\
PromptRoBERTa &83.82& 88.72& 93.19& \textbf{90.36}& 88.08& 90.60& 76.75& 87.36\\
RankCSE&83.32&  88.61&  94.03&  88.88&  89.07&  \textbf{90.80}&  76.46&  87.31\\
\hline

MultiCSR (SimCSE)&\textbf{84.70}	&  \textbf{90.69}&  	\textbf{94.40}&  	89.38&  	\textbf{89.42}&  	89.62&  	\textbf{77.01}&  	\textbf{87.89}\\

\hline
\end{tabular}
\vspace{-3mm}
\caption{Performance comparison on transfer tasks. *: since they select the model based on the development sets of transfer tasks, we retest their performance with their officially released checkpoints for a fair comparison.
} \label{tab:transfer}
\vspace{-4mm}
\end{table*}

\begin{table}[!ht]
\centering
\renewcommand\arraystretch{1.1}
\setlength{\tabcolsep}{11pt}
\begin{tabular}{llc}
\hline
\textbf{$L$}&\textbf{$\lambda$} &\textbf{Spearman’s}\\
\hline
\hline
\multicolumn{3}{c}{\textit{BERT-base}} \\
\hline
3&0.6  & 84.62 \\
5&0.8 & 84.71 \\
\hline
\ \ - & \ \ - & 84.55\\
\hline
\hline
\multicolumn{3}{c}{\textit{RoBERTa-base}} \\
\hline
3&0.6  & 85.72 \\
5&0.8 & 86.01 \\
\hline
\ \ - & \ \ - & 85.83\\

\hline
\end{tabular}
\vspace{-3mm}
\caption{Studies of different number of demonstrations $L$ and similarity controlled thresholds $\lambda$ in MultiCSR. The results are the development performance of STS-B.
} \label{tab:demos}
\vspace{-3mm}
\end{table}

\section{Supervised Settings} \label{appendix_demons}
In this section, we introduce how we perform supervised settings on our method.
Using demonstrations is now a standard way to perform few-shot~\cite{NEURIPS2020_1457c0d6, Agrawal2022IncontextES} inference on LLMs in various tasks. As a natural semantic retriever, the pre-trained sentence representation model $P_{\eta}$ can be utilized to find the most proper demonstrations. Given the source sequence $\boldsymbol{x}$, we first compute its representation as $P_{\eta}(\boldsymbol{x})$, and search over $\mathcal{W}$ (i.e., NLI premises specifically) to find the most relevant demonstrations, based on ${\rm{sim}}(P_{\eta}(\boldsymbol{x}), P_{\eta}(\boldsymbol{x}'))$ where ${\rm{sim}(,)}$ calculates the cosine similarity of two parameterized vectors. We denote the set of $L$ demonstrations as $\mathcal{D} = (d_1, d_2, ..., d_L)$, and each demonstration $d$ will be either the entailment or contradiction hypothesis of the premise $\boldsymbol{x}^k$. To prevent the low similarity demonstrations, we only choose premise $\boldsymbol{x}^k$ with ${\rm{sim}}(P_{\eta}(\boldsymbol{x}), P_{\eta}(\boldsymbol{x}^{k})) \geq \lambda$, where $\lambda$ serves as a hyperparameter threshold. Finally, the whole text generation procedure without any parameter updates can be formulated as the following:
\begin{equation*} \label{eq:5}
\boldsymbol{y} \gets C_{\theta}\left([\mathcal{D}; \boldsymbol{x}; I]\right),\, \boldsymbol{x} \in \mathcal{W}.
\end{equation*}

We have included the performance of several combinations of different $L$ and $\lambda$ in Table \ref{tab:demos}. From the result, we can see that, although using demonstrations can be still helpful in our contrastive generation procedure, the improvement compared with not using demonstrations is not so significant. And inference with more demonstrations will somehow increase the time required for generation. Thus, in all our supervised settings, we choose not to use demonstrations including the results in Table \ref{tab:sup_ours}.

\begin{figure}
  \centering
  \includegraphics[width= 0.9\linewidth]{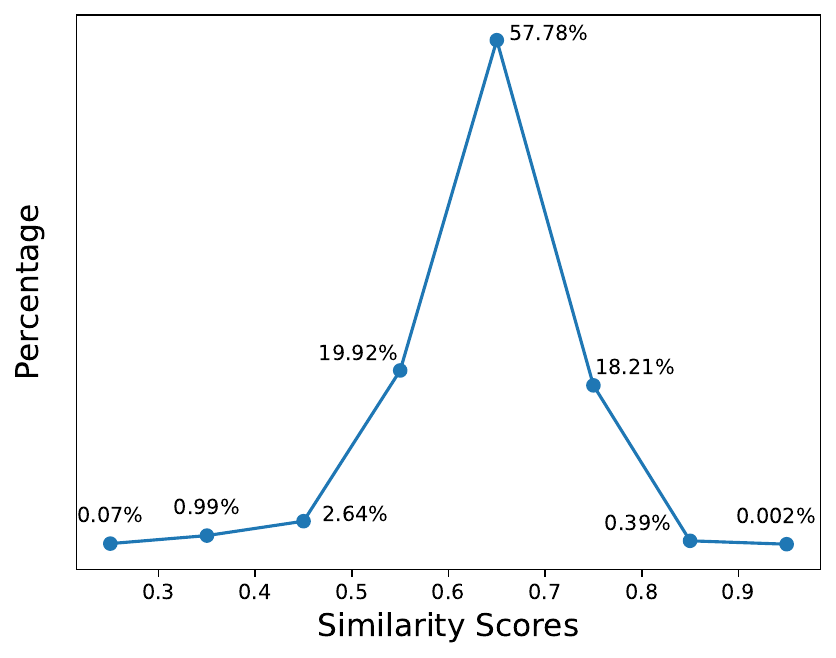}
  \vspace{-4mm}
  \caption{The percentages of consine similarity scores provided by pre-trained supervised SimCSE-RoBERTa on a randomly sampled batch of size 256.}
  \label{fig:statistic}
  \vspace{-4mm}
\end{figure}

\begin{table}[t]
\centering
\renewcommand\arraystretch{1.1}
\small
\setlength{\tabcolsep}{13pt}
\begin{tabular}{lc}
\hline
\textbf{Model} &\textbf{Avg.}\\
\hline
\hline
\multicolumn{2}{c}{\textit{BERT-base}} \\
\hline
SBERT & 74.89 \\
SimCSE  & 81.57 \\
\hline 
MultiCSR w/o stage3 & 81.62 \\
MultiCSR & 81.97 \\
\hline
\hline

\multicolumn{2}{c}{\textit{RoBERTa-base}} \\
\hline
SRoBERTa  & 74.21 \\
SimCSE  & 82.52 \\
\hline
MultiCSR w/o stage3 & 82.27 \\
MultiCSR & 82.70 \\
\hline
\end{tabular}
\vspace{-2mm}
\caption{Supervised performance on STS tasks.
} \label{tab:sup_ours}
\vspace{-6mm}
\end{table}

After generation of $\mathcal{T}$, we would combine $\mathcal{T}$ with ground truth corpus $\mathcal{N}$, which means, for each premise in $\mathcal{T}$, there would be at least one triplet with the same premise in $\mathcal{N}$. As an example, we randomly sample a batch of size $N$ from $\mathcal{T}\cup \mathcal{N}$, the distribution of cosine similarity given by ${\rm{sim}}(P_{\eta}(\boldsymbol{x}), P_{\eta}(\boldsymbol{x}'))$ between any premises and $(h^k_+,h^k_-)$ (i.e., $2N(N-1)$ similarity scores here) is shown in Figure \ref{fig:statistic}. From the figure, we can see that even excluding the entailment and contradiction hypotheses of each premise, there are still more than 18.6\% sentences inside a batch having a similarity score higher than 0.7. Thanks to mask indicator introduced in Section \ref{sec:loss}, we can directly utilize this mixed corpus for training without suffering from \textit{false-negative} problem. The results of supervised MultiCSR are included in Table \ref{tab:sup_ours}.

\section{Cost Analysis} \label{appendix_cost_analysis}
We further provide a cost analysis of the time and space. Since our stage 1\&2 works before the in-batch training stage, we provide the analysis of the resource required to support our stage 3, under different batch sizes. We test on a single NVIDIA V100-32GB, and replicate a triplet $10K$ times as the training corpus. The time and GPU memory required for training are shown in Table \ref{tab:time}, while the inference time and space of our method remains the same as theirs.

\begin{table*}[]
\centering
\renewcommand\arraystretch{1.1}
\small
\setlength{\tabcolsep}{5pt}
\begin{tabular}{llc}
\hline
\textbf{Label} &\textbf{Original \& Generated Sentences}&\textbf{Semantic Similarities}\\
\hline

\textbf{Premise} & One of our number will carry out your instructions minutely. & -\\
\textbf{Entailment} & A member of my team will execute your orders with immense precision. & 4.5\\
\textbf{Contradiction} & We have no one free at the moment so you have to take action yourself. & 0.0 \\
\hline
\textbf{Premise} & He turned and smiled at Vrenna. & -\\
\textbf{Entailment} & He turned back and smiled at Vrenna. & 5.0\\
\textbf{Contradiction} & He turned and walked away. & 0.0 \\
\hline
\textbf{Premise} & How do we fix this? & -\\
\textbf{Entailment} & How can we fix this? & 5.0\\
\textbf{Contradiction} & Let's not worry about fixing this. &{1.0} \\
\textbf{w/o \textcolor{stage1}{Stage 1}} & We can't figure out how to fix this. & \textcolor{red}{4.0} \\
\hline
\textbf{Premise} & The economy could be still better. & -\\
\textbf{Entailment} & The economy is not at its best possible state. & 4.0\\
\textbf{w/o \textcolor{stage1}{Stage 1}} & The economy is not good. & \textcolor{red}{0.0}\\
\textbf{Contradiction} & The economy could be worse. & 0.0\\
\hline
\end{tabular}
\vspace{-2mm}
\caption{Generated sentence triplets and the semantic similarities between the hypotheses and their premises. The sentence and similarities are generated by  on NLI premises. If we set the thresholds of filtering strategy as $\alpha=3$, $\beta=3$ and $\gamma=1$, the third triplet will not appear in the training corpus $\mathcal{T}$ because of unqualified contradiction $4.0\nleq3$, and neither for the fourth triplet because of wrong entailment $0.0\ngeq3$.
} \label{tab:example}
\vspace{-3mm}
\end{table*}

\begin{table}[]
\centering
\small
\setlength{\tabcolsep}{2.5pt}
\begin{tabular}{c|cc|cc}
\toprule
\multicolumn{1}{c|}{\multirow{2}{*}{\textbf{Batch Size}}} & \multicolumn{2}{c|}{\textbf{Time(s)}} & \multicolumn{2}{c}{\textbf{GPU Memory(MB)}}\\ 
\cmidrule(lr){2-3} \cmidrule(lr){4-5}
&\textbf{SimCSE} & \textbf{MultiCSR} &\textbf{SimCSE} & \textbf{MultiCSR}  \\
\hline
\textbf{8 }         & 86(1x)      & 99(1.15x) & 3863(1x)         & 5225(1.35x)    \\ \hline
\textbf{16}         & 62(1x)      & 81(1.31x) & 4873(1x)         & 6080(1.25x)    \\ \hline
\textbf{32}         & 55(1x)      & 75(1.36x) & 5661(1x)         & 8163(1.44x)    \\ \hline
\textbf{64}         & 52(1x)      & 68(1.30x) & 7353(1x)         & 11965(1.63x)   \\ \hline
\textbf{128}        & 51(1x)      & 65(1.27x) & 11503(1x)        & 20137(1.75x) \\
\hline

\end{tabular}
\vspace{-2mm}
\caption{Time and memory cost analysis of perform our stage 3, contrastive in-batch training.
} \label{tab:time}
\vspace{-4mm}
\end{table}

It is worth mentioning that the data generated in the first stage of our framework becomes more lightweight after the second stage of self-curation. This makes, although the time required to train on the same dataset increases, the time we need to train a model of MultiCSR is actually greatly reduced. We assume that a training set has $M$ sentences, and the batch size is $N$. In unsupervised SimCSE, for each sentence, it will have $1$ positive pair and $N-1$ negative pairs, then there will be a total of $M*N$ pairs of similarity to be calculated during training. In MultiCSR, for each sentence, it will have $1$ positive pair and $2N-1$ negative pairs, a total of $M*(2N)$ pairs will be considered. But in our scenario, the value of $M$ varies greatly. For example, if we set the batch size to 64, it takes $0.98h$ to train an unsupervised SimCSE on $1M$ Wikipedia sentences, with $1M*64 =64M$ pairs. But after our first and second stages with these sentences, only $0.19M$ triplets are left, with $0.19M*(2*64) = 24.32M$ pairs. Based on this, it only takes $0.36h$ to train our model.

\section{Case Studies} \label{appendix_example}
In this section, we present some generated triplets using Flan-T5-XL in Table \ref{tab:example} with the premises from NLI and semantic similarity scores for these triplets. While the first two example triplets are high quality data for training, the last two triplets suffer from either the \textit{false-negative} or the \textit{false-positive} problem if Stage 1 is not applied. For example, in the third triplet, the contradiction sentence generated without Stage 1 ``We can't figure out how to fix this.'' has a 4.0 high similarity score to the premise sentence ``How do we fix this$?$'', indicating a \textit{false negative} that can harm the training if included. In our method, we can either avoid these ``low-quality'' triplets by introducing Stage 1 or self-curation strategy by $\alpha=3$, $\beta=3$ and $\gamma=1$, which we proposed in Stage 2. For another example, we can also exclude the third triplet in training due to unqualified contradiction $4.0\nleq3$ and avoid the \textit{false negative} problem. From the case studies we conduct, we can see that all components are helpful in refining the generation of LLMs, ensuring only high-quality sentence pairs are utilized into model training.

\end{document}